# Training of Neural Networks with Uncertain Data – A Mixture of Experts Approach

**Lucas Luttner**

Faculty for Computer Science and Data Science, University of Regensburg

Lucas.Luttner@ur.de

**Abstract**

This paper introduces the "Uncertainty-aware Mixture of Experts" (uMoE), a pioneering solution aimed at addressing aleatoric uncertainty within Neural Network (NN) based predictive models. While existing methodologies primarily concentrate on managing uncertainty during inference, uMoE uniquely embeds uncertainty into the training phase. Employing a "Divide and Conquer" strategy, uMoE strategically partitions the uncertain input space into more manageable subspaces. It comprises Expert components, individually trained on their respective subspace uncertainties. Overarching the Experts, a Gating Unit, leveraging additional information regarding the distribution of uncertain inputs across these subspaces, dynamically adjusts the weighting to minimize deviations from ground truth. Our findings demonstrate the superior performance of uMoE over baseline methods in effectively managing data uncertainty. Furthermore, through a comprehensive robustness analysis, we showcase its adaptability to varying uncertainty levels and propose optimal threshold parameters. This innovative approach boasts broad applicability across diverse data-driven domains, including but not limited to biomedical signal processing, autonomous driving, and production quality control. Code can be accessed here: https://github.com/lucascodinglab/uMoE_Training_Neural_Networks_with_uncertain_data

## Introduction

Data uncertainty is a prevalent challenge within the domain of Deep Learning, presenting significant obstacles in the development and application of such predictive models. It is a challenge that permeates various domains, from medicine (Raviv and Intrator 1996) to autonomous driving (Meyer and Thakurdesai 2020) and sensors-based domains like quality control in manufacturing (Rodríguez and Servigne 2013) casting a shadow of ambiguity over the reliability of data-driven decisions (Hariri et al. 2019).

Addressing this uncertainty is paramount for the effective training of Deep Learning models. NNs, a cornerstone of contemporary Deep Learning, thrives on deterministic data (Loquercio et al. 2020). However, when confronted with the real-world complexity of uncertain data, these models falter (Ge et al. 2010). It is not merely a matter of accuracy; the stakes are higher. In applications like autonomous driving, misinterpretation of uncertainty can have dire consequences (Feng et al. 2018). Especially in cases, where only uncertain data is available, considering this additional information during training is essential, rather than just during inference. However, in the existing research on uncertainty in Deep Learning, the primary focus is predominantly on the propagation of uncertainty through NNs during inference. However, these methods assume that the model used for inference has been trained on certain data without considering uncertainty in the input data, which can be represented in form of Probability Density Functions (PDFs) over the uncertain attributes. Due to this limitation, these methods are not suitable for cases where the input space already exhibits uncertainty (Astudillo and Neto 2011; Abdelaziz et al. 2015; Smieja et al. 2019).

Apart from this major research path, the already limited number of existing approaches in the literature regarding the training of NNs under uncertainty is subject to many restrictions. These restrictions result for example from the assumption of parametric PDFs, from which they use properties of the distribution, such as mean value and variance, for considering uncertainty during training. This is a problem in cases, where uncertainty is not following a specific distribution. Besides that, these methods are just applicable for specific domains such computer vision or probabilistic inference and are thus not addressing uncertainty as their main research problem (Kendall and Gal 2017; Gast and Roth 2018). Furthermore, such approaches tend to focus exclusively on uncertainty in the labels rather than questioning the underlying attribute uncertainty (Nguyen et al. 2019; Huang et al. 2022).

Derived from the significance of the problem and the research gap in the existing literature on uncertain data, the central research question guiding this work can be formulated as:

*"How can a NN-based predictive model be trained under the presence of aleatoric uncertainty in the input features?*

To address this question comprehensively, this paper introduces uMoE, an innovative approach that addresses the challenge of training NN-based predictive models in the presence of aleatoric uncertainty, while being able to work with any type of PDF in the input space. It stands as a pioneering technique in the landscape of Deep Learning, of-fering a novel way to adapt NNs to uncertainty. Unlike traditional methods that rely on predetermined analytical formulas constrained by distribution types, the uMoE method leverages a Gating Unit that flexibly responds to different uncertainty distributions, adapting to and learning from uncertainty as the final decision-maker.

In the following sections, we will discuss categories and characteristics of uncertain data, highlighting the distinction between aleatoric and epistemic uncertainty. We will examine the origins of the Mixture of Experts (MoE) framework, its core principles, and its role as a precursor to our uMoE method. Additionally, we will provide a detailed overview of the uMoE architecture and explain how it handles uncertainty. We will present comprehensive evaluations demonstrating uMoE's superior performance compared to baseline NN models in the presence of uncertain training data. This journey into uncertainty-aware Machine Learning underscores the importance of addressing uncertainty within modelling of NNs to enhance the robustness and reliability of AI systems in an increasingly complex and uncertain world.

## Related Work

In this chapter, we review related work to explore existing approaches for handling uncertain data during the training of predictive models. It is worth noting that, the number of papers confronting this problem is already manageable little and underlies several limitations. Such limitations are related to the type of probability functions the approach can handle, mostly speaking of parametric distributions like Gaussian, their applicability to various prediction tasks, or specific application domains like computer vision. Existing literature can be broadly categorized into three groups. The first group addresses uncertainty in input features, similar to our approach, although not within the context of NNs. The next section, closely related to our work, presents the literature that considers this uncertainty within NNs. Finally, we shortly introduce approaches that, while addressing uncertainty in the training dataset, focus primarily on labels instead of PDFs over features and are therefore only mentioned in passing.

We begin with approaches, that focus on training of non NN-based models under uncertainty in the input features. In one such approach, the authors address the challenge by extending the Naïve Bayes algorithm. They propose three extensions: one based on averaging, which computes expected values and uses them as input for predictions, a sample-based method that considers values sampled from PDFs, and a formula-based method designed for specific parametric probability distributions e.g. Gaussian distributions (Ren et al. 2009).

In a second approach, similar to the first, a classifier model is adapted to handle uncertain data as PDFs. For this purpose, the authors used a classic decision tree algorithm as framework and modified it in a way, that it divides the PDF into probability intervals for each split point and incorporates these weighted intervals into the entropy loss during training (Tsang et al. 2011).

In addition to these two approaches, there are also more related methods, that not only consider uncertainty in the training process, but also apply it in the context of NNs.

Gast and Roth (2018) address the propagation of uncertainty through NNs, emphasizing a method that prioritizes fast runtime with minimal adjustments to the NN architecture. Among other techniques, they implement a probabilistic output layer and replace intermediate activation functions with assumed density filtering. In the process of propagating uncertainty, they also consider the uncertainty in the data during the training of such an NN architecture. The primary focus is on maximum conditional likelihood learning, which seeks to maximize the conditional likelihood of data under a predictive model.

From another perspective, Kendall and Gal (2017) developed a framework for Bayesian Deep Learning that allows modelling of input-dependent aleatoric uncertainty alongside epistemic uncertainty. The framework was developed for computer vision tasks such as semantic segmentation and depth regression. Concerning the training of NNs, they use a loss function that incorporates aleatoric uncertainty into the modelling of predictions. The authors adapt the loss function for regression tasks, incorporating aleatoric uncertainty by weighing data points based on predicted variance. For classification tasks, they introduce a Gaussian distribution over logits, creating a stochastic loss function that encourages the model to learn loss attenuation based on uncertainty. However, it is important to note that their modelling of aleatoric uncertainty is limited to a single dispersion parameter, leaving many other properties of the PDF unused.

Although the four approaches described share similarities with our goal of training NN-based predictive models with uncertain data, each method exhibits distinct limitations. Ren et al.'s method is exclusively suitable for classification tasks and is restricted to Naïve Bayes, which relies on the specific assumption, that all features are independent. These limitations also apply to the second approach, which is as well only suitable for classification tasks and just works in the context of decision trees. While Gast and Roth employ a NN model, their approach extends beyond training with uncertain data, and it still assumes parametric probability distributions, not considering data to be certain after training. In this method, training serves primarily to enhance the model's robustness for inference under uncertainty. Finally, Kendall and Gal's approach, despite being based on a Bayesian-NN architecture, predominantly assumes parametrically distributed noise over the data. Moreover, this method is primarily designed for computer vision applications, where the input space is significantly larger and not easily transferable to tabular data due to the intricacies introduced by convolutional layers.

Apart from the mentioned approaches to consider uncertainty in the input data during training, there are also methods that address uncertainties in the target variable. It is important to note that these methods, while dealing with aleatoric uncertainty in the training process, differentiate themselves from our idea by exclusively focusing on noise or uncertainty in the labels. Therefore, they are mentioned here only in passing. In one approach, the authors tackle label noise in NNs, particularly in imbalanced datasets. They present the Uncertainty-aware Label Correction framework, which consists of two key components: one for modelling epistemic noise to manage variations in class-specific loss distributions, and another for aleatoric uncertainty-aware learning to address residual noise (Huang et al. 2022). In the second approach the authors address label noise in Machine Learning models more broadly. They propose iterative filtering with Semi-Supervised Learning, an approach that iteratively removes noisy labels while retaining associated data samples. This is accomplished through an unsupervised loss term acting as a regularization to combat label noise (Nguyen et al. 2019).

# Background

Now that we have addressed the context of our work, as well as the related methods, we take a closer look at the background of our uMoE. Firstly, we present an overview of data uncertainty to better understand the problems that arise with it. We then establish a fundamental understanding of the MoE framework in which we have discovered the potential to use it for training of predictive models with PDFs.

**Categories and Characteristics of Uncertain Data**

In the realm of data-driven decision-making and Machine Learning, data is often assumed to be precise and deterministic. However, in the real world, data is frequently imbued with various forms of uncertainty. This uncertainty can emanate from multiple sources and can significantly impact the reliability and robustness of data-driven models. Uncertain data can be broadly categorized into two main types (Murphy 2012): aleatoric and epistemic uncertainty. Aleatoric uncertainty arises from the stochastic or random nature of the underlying processes generating the data, representing irreducible randomness in observations. It can be observed in scenarios such as sensor noise in measurements (Wang et al. 2013), natural variability in medical data (Alizadehsani et al. 2021) or fluctuations in financial markets (Li et al. 2020). Epistemic uncertainty, on the other hand, is related to the lack of a model to find the optimal parameters, particularly in the context of NNs, where it arises from the randomness of the stochastic gradient descent, resulting potentially in a local minimum (Helton et al. 2010; Jiang et al. 2018). In our case, we are addressing the issue of aleatoric uncertainty in input data while assuming given labels are certain. This means that instead of assuming attributes as certain scalars, we consider them to be represented by multidimensional continuous PDFs over the uncertain attributes. These PDFs can take any form of distribution, such as Gaussian, Poisson, or even non-parametric distributions. By modeling the uncertain attributes with PDFs, it allows our method to capture the inherent variability and uncertainty in the data (Hora 1996).

**Introduction to Mixture of Experts**

The MoE framework represents a significant advancement in the field of Machine Learning and ensemble methods (Gormley and Frühwirth-Schnatter 2018). Its origins trace back to 1991 when Jacobs et al. introduced the concept of decomposition theory in ensemble learning, notably the "Divide and Conquer" paradigm. This paradigm entails a structured approach to solve complex problems. It involves breaking down intricate challenges into smaller, more manageable subproblems. These subproblems are then addressed independently, and their solutions are eventually combined to derive the solution for the original problem (Jacobs et al. 1991). This principle comprises three fundamental steps (Smith 1985):

1. **Divide**: Initially, the overarching problem is partitioned into smaller instances of the same problem or entirely new subproblems. Often, this division process is recursive, meaning each subproblem may undergo further division until they become straightforward to solve.

2. **Conquer:** In this phase, the divided subproblems are individually solved. This can entail applying the same algorithm recursively or employing diverse methods based on the nature of each subproblem. The primary goal is to find effective solutions for these subproblems.

3. **Combine:** After solving the subproblems, their solutions are integrated or combined to arrive at a solution for the original, more complex problem. This combining step is crucial as it ensures that the solution aligns correctly with the original problem.

At its core, the MoE framework is firmly grounded in this paradigm. It strategically deconstructs a complex problem (for example the entire data space $x \in \mathbb{R}^d$, where $d$ represents the input dimensionality) into more manageable, smaller subspaces.

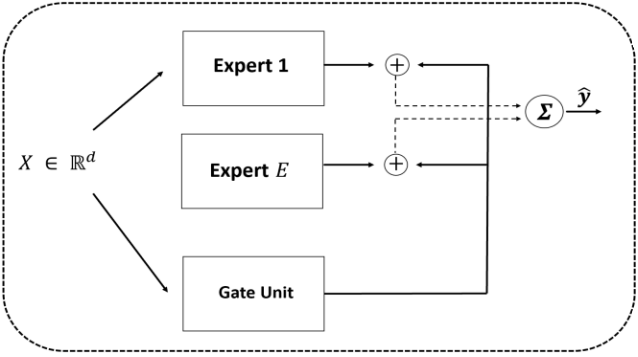

**Figure 1: Illustration of the Mixture of Experts framework**

The MoE architecture primarily comprises two key components, which are responsible for solving the three steps: Expert components and a Gating Unit. Experts can be likened to specialized problem solvers, with each Expert dedicated to a specific segment of the problem domain. In general, almost any conceivable predictive Machine Learning method can serve as an Expert, provided it aligns with the objectives and complexity of the prediction task. To allocate subgroups of the problem to Experts, a common approach involves employing a cluster-based unsupervised Machine Learning method. This method partitions input data based on shared data distributions among attributes, akin to the Divide step. Once the problem is subdivided, an Expert is assigned to each subspace. Following this, each Expert is assigned all instances that are located within its region. The amount of all instances per Expert constitutes the training dataset. This avoids including instances that are outside of the Expert's region, thus minimizing the need for generalization while finding the optimal parameters. This process aligns with the Conquer step, as the problem is effectively conquered within each subgroup. Conversely, the Gating Unit takes on the role of distributing input data among these Experts. Drawing inspiration from the Combine step, it learns to identify which Experts are best equipped to handle different types of inputs. It then uses weighted decisions based on the predictions from each Expert to allocate inputs to the most suitable ones (Yuksel et al. 2012; Masoudnia and Ebrahimpour 2014). Mathematically formulated: The prediction for a given instance is $\hat{y} \in \mathbb{R}$, which is calculated as the weighted sum of the output predictions from each Expert $\hat{y}_e$ multiplied by the corresponding weight of the Gating Unit $g_e$:

$$\hat{y} = \sum_{e=1}^{E} g_e \hat{y}_e$$

*Note that: $E \in \mathbb{N}$ represents the number of Experts.*

The advantages of the MoE concept are manifold. Firstly, it significantly enhances efficiency by breaking down complex problems into smaller, more manageable subproblems, resulting in improved overall performance. Secondly, MoE exhibits remarkable flexibility and adaptability, accommodating various types of Experts and problem domains (Maimon and Rokach 2005). Building upon the foundational principles of MoEs, our uMoE approach emerges. From our standpoint, we perceive aleatoric uncertainty as an additional layer of complexity situated behind the decomposition of the input space. In this light, our objective is twofold. First, it entails incorporating uncertainty into the process of decomposing the input space. Second, it involves harnessing the insights gained from this decomposition under conditions of uncertainty. These insights are then translated into additional information and presented to the Gating Unit as supplementary information. These methodological steps clearly demonstrate why MoEs have significant potential when dealing with uncertainty. In contrast to most approaches in literature, uncertainty is not incorporated into a purely analytical formula that considers parameters such as the expected value and variance of uncertainty (and thus is subject to various constraints, such as the distribution type). Instead, this process is controlled by a Gating Unit, which can flexibly respond to different distributions and, as a result, adapt to and learn from the uncertainty as final decision maker. This flexibility is particularly crucial in areas where either the nature of uncertainty cannot be predetermined in advance or where a Gaussian distribution is not appropriate.

## Uncertainty-aware Mixture of Experts Model

In this chapter, we present our method – Uncertainty-aware Mixture of Experts (uMoE) − for training a NN-based model with uncertain data, represented by PDFs of any shape. Although aleatoric uncertainty in the input space is often prevalent in many real-world scenarios, standard Deep Learning algorithms solely can operate with certain data points, ignoring the variance of the uncertainty. To address this, our method aims to train a NN-based model by separating the uncertainty into its constituent subspaces and leveraging the information derived from this decomposition. Firstly, we establish the mathematical problem and thus the relevance for our method, the definition of an uncertain instance. Next, we introduce NNs as predictive models within our method for Experts, as well as for the Gating Unit. After presenting the backbone of our method, the following five sections introduce the concept of our uMoE for partitioning the uncertain input space among the individual Experts responsible for this subspace. After the description of our method, we then further define the Nested Cross-Validation (NCV), as process to find the optimal number of subspaces for the uMoE.

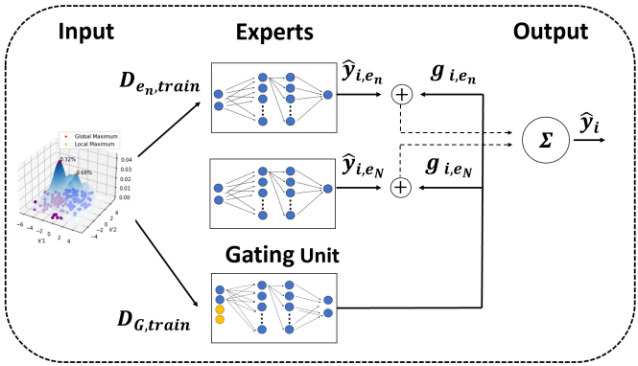

**Figure 2: Overview of the training process of uMoE**

Figure 2 serves as a fundamental overview of the uMoE (here illustrated with two Experts). The model's initiation entails considering an uncertain instance, which is subsequently allocated to an Expert, by partitioning the input space to identify the region where the PDF exhibits the highest probability mass (see section 1). Within the subspace, the local mode value is calculated, which is defined as the maximum of a PDF within the corresponding subspace. This process is iteratively applied to each instance until all Experts have been trained using the local mode values, which they are responsible for as described in, section 2 and 3. Subsequently, the Gating Unit is trained on the global maximum of the PDFs, which additionally incorporates the distribution of the PDF over the subspaces as additional information (see section 4). It learns to weigh the predictions of the Experts during this process in a manner that minimizes the deviation of model-predictions from the ground truth.

**Definition 1: Uncertain instance**
This section serves as an introduction to the problem of aleatoric uncertainty in the input space. Let $D_{train}$ be the dataset used for training the uMoE model. Let $D_{train}$ contain $I$ instances with $k$ attributes. Each instance $d_i \in D, i \in \{1, ..., I\}$ can be represented as:

$$d_i = (x_{i1}, x_{i2}, ..., x_{Ik})$$

with an affiliated PDF denoted as:

$$pdf_i: \mathbb{R}^{k'} \to \mathbb{R}^+, k' \leq k$$

*Note that: $k'$ represents the set of attributes $k$ of an instance that are uncertain. The definition of an uncertain instance is applicable to all continuous attributes, as only they can be meaningfully transformed into a continuous function such as in our context a PDF. The attributes of an instance, which are not affected by uncertainty are defined as $x_{ik*}$, where $k* \subseteq k$.*

This PDF $pdf_i$ characterizes the uncertainty associated with the instance, mapping the $k'$-dimensional attribute space to $\mathbb{R}^+$. In the context of our model, these PDFs encapsulate the uncertainty inherent in each data point. Importantly, our method does not impose restrictive assumptions regarding the distribution of uncertainty, such as assuming a Gaussian distribution. Instead, it offers the flexibility to directly incorporate any type of continuous distribution into the model. This flexibility is particularly valuable, as it allows us to tailor our approach to a wide range of applications, where uncertainty may manifest in diverse and complex ways.

**Definition 2: Neural Networks as predictive models within uMoE**
Following the introduction of uncertain data in definition 1, this definition explores another fundamental aspect of the uMoE architecture. In the background chapter, Experts and the Gating Unit were initially viewed as placeholders for various types of Machine Learning models. In this section, we specifically define NNs as predictive models within our architecture. Assuming we have for each Experts $E = \{e_{n=1}, e_{n=2}, ..., e_N\}$ and Gating Unit $G$ one NN with $L$ layers and $T$ neurons in each layer. A special feature of the Gating Unit compared to the NN architecture of the Experts is, that the output is always defined as softmax function, where the number of neurons in the output layer equals the number of Experts. Apart from this, output of the $t$-th neuron $\{t \in 1, 2, ..., T\}$ in the $l$-th hidden layer $\{l \in 1, 2, ..., L\}$ can be represented as:

$$z_t^{(l)} = \sum_{t=1}^{T} w_{t,t-1}^{(l)} a_{t-1}^{(l-1)} + b_t^{(l)}$$
$$a_t^{(l)} = f(z_t^{(l)})$$

<u>where</u>:

- $z_t^{(l)}$ is the weighted sum of inputs for the $t$-th neuron in the $l$-th layer
- $w_{t,j}^{(l)}$ is the weight between the $t-1$-th neuron in the previous layer and the $t$-th neuron in the current layer
- $b_t^{(l)}$ is the bias term of the $t$-th neuron in the $l$-th layer
- $a_t^{(l)}$ is the output of the $t$-th neuron after applying the activation function $f: \mathbb{R} \to \mathbb{R}$ on $z_t^{(l)}$

**Section 1: Cluster-based decomposition of probability density functions**
This section delves into the core idea of the concept behind uMoE, which involves simplifying the intricate issue of uncertainty by breaking it down into its constituent parts. To achieve this, we make use of the concept of sampling from a PDF to further divide the PDF through cluster-based algorithms.

To decompose the input space into subspaces, a set of $M$ samples per instance is drawn from $pdf_i$ to get $S_i = \{s_{i,1}, s_{i,2}, \ldots, s_{i,M}\}, s \in \mathbb{R}^{k'}$. Next, the threshold parameter $p \in (0,1]$ is introduced in this context to reduce the sample selection by filtering out $p$ percent of the samples with the lowest density. This threshold allows our method to react to a different amount of variance in the PDFs through limiting the region from which the samples are drawn.

To achieve the threshold sampling, let the density of a sample point $s_{i,m}$ be determined as $pdf_i(s_{i,m})$. The density of a specific point in a PDF represents the likelihood of that point occurring within a continuous probability distribution. Next, we order the PDF values of the samples as a sequence $\rho_i = (\rho_{i,1}, \ldots, \rho_{i,M})$ with $\rho_{i,j} \geq \rho_{i,j+1}$ for all $1 \leq j \leq M-1$ and $\rho_{i,j} = pdf(s_{i,\sigma(j')})$ for a (bijective) permutation $\sigma: \{1, \ldots, M\} \to \{1, \ldots, M\}$. To obtain the $p$ share of highest PDF values, we define the subsequence $\rho_{i,p} = (\rho_{i,1}, \ldots \rho_{i,M_p})$ consisting of the first (and thus largest) $M_p = \lceil p \cdot M \rceil$ values of $\rho_i$. This subsequence corresponds to a subset $S_{i,p} \subset S_i$ via $\sigma$:

$$S_{i,p} = \{s_{i,j} \in S_i | \sigma(j') = j, 1 \leq j' \leq M_p\}$$

In other words, $S_{i,p}$ contains the samples with the $p$ highest share of PDF values.

This process of sorting out samples with low probability allows our method to react to different amount of uncertainty in the data and thus to make our method more robust. After sample reduction, the next step is to decompose the uncertainty of an instance in different subspaces corresponding to each Expert via a clustering procedure.

To decompose $D_{train}$ into subspaces, the k-means algorithm with $|E|$ clusters is applied on all samples $S_{i,p}$, where $|E|$ represents the total number of Experts. This results in a model predicting the cluster affiliation of the samples $S_{i,p}$:

$$kmeans(S_{i,p}): \mathbb{R}^{k'} \to [1, 2, \ldots, |E|]$$

We calculate the cluster affiliation of instance $i$ based on the relative frequencies in $kmeans(S_{i,p})$ and obtain a vector of cluster probabilities:

$$C_i = [c_{i,j=1}, \ldots, c_{i,|E|}], c_{i,j} \in [0,1] \, \forall i, j$$

This formula can be described as the probability $c_{i,j}$ of instance $i$ to belong to cluster $j$ or in other words the relative distribution of the samples $S_{i,p}$ across the subspaces.

**Section 2: Determination of global and local mode value**
Having now divided the input space, the next step is to determine the global maximum of the PDF, as well as the local maximum within the corresponding subspace of the Experts. The local mode value is subsequently used for training the Experts, and the global mode value serves for training the Gating Unit. Using a global optimizer, the global mode $m_{i,global} \in \mathbb{R}^{k'}$ of the PDF over the uncertain attributes of an instance can be computed as:

$$m_{i,global} = argmax \, (pdf_i(\dot{x}))$$

*Note that: $\dot{x}$ is the input configuration that undergoes optimization, with the to find the point, that maximizes the PDF.*

For Expert $e_n$, as the whole input space of $pdf_i$ is decomposed into subspaces through section 1, the aim is to uphold the assumption of MoEs, that an instance is assigned to the Expert responsible for the subspace it belongs to.

To find the local mode value $m_{i,local} \in \mathbb{R}^{k'}$ for every instance $i$ containing uncertain attributes $k'$, the maximum of the cluster probability Vector $C_i$ is calculated.

$$C_{i,max} = argmax \, (C_i)$$

$C_{i,max}$ determines in which cluster $c_{i,j}$ most of the probability mass defined through $S_{i,p}$ of instance $i$ lies, which will serve on the one hand to restrict the search space of the local mode value within $pdf_i$ and on the other as weighting factor for the loss function during training of the corresponding Expert (see section 3). The subspace-restricted maximization algorithm can thus be defined as:

$$m_{i,local} = argmax \, (pdf_i(\dot{x}))$$

*subject to*:

$$\dot{x} \in C_{i,max}$$

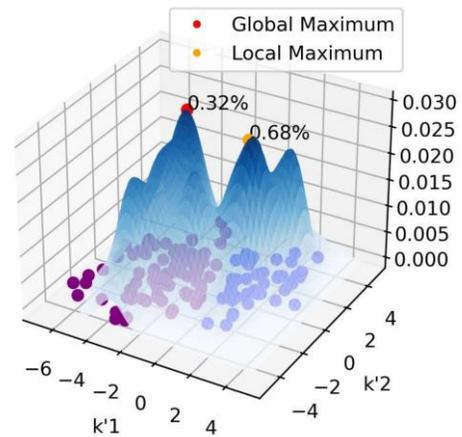

Figure 3: Determining the local mode value in two-dimensional space with two subspaces

Figure 3 exemplifies the determination of the $m_{i,local}$. Here, the uncertain instance $i$ can be expressed as $pdf_i: \mathbb{R}^2 \to \mathbb{R}_0^+$. The global mode $(k'1 = -4, k'2 = 2)$ value lies outside the Cluster $C_i[C_{i,max}] = 0.68$ where most samples $S_{i,p}$ are located. Therefore, within dominant Cluster, the local mode value $(k'1 = -1, k'2 = 3)$ is determined restrictively. This value is subsequently assigned to the Expert $e_n$ responsible for this subspace for training purposes.

**Section 3: Training of Experts with weighted loss**
By decomposing the PDF, each Expert can now be assigned the local mode value for training, which resides within its respective region. Let $D_{experts,train}$ be the training dataset for the Experts, includes instance $x_{ik}$, which includes the local mode values $m_{i,local}$ for the uncertain attributes of the instance $i$, as well as the certain attributes $x_{ik*}$. The instances are finally allocated as trainings dataset $D_{e_n,train} \subseteq D_{experts,train}$ to every Expert by the cluster object $kmeans(x_{ik})$ from section 1 under the restriction, that the instance lies within the subspace of $e_n$ (see section 2):

$$D_{e_n,train} = \{(m_{i,local}, x_{ik*}) | kmeans(m_{i,local}, x_{ik*}) \in e_n\}$$

The training process can be formulated as an optimization problem, where the parameters $\theta = (w_{t,t-1}^{(l)}, b_t^{(l)})$ of the Expert are adjusted in such a way that the difference between the predicted output of the NN $\hat{y}_i \in \mathbb{R}$ and the actual output $y_i \in \mathbb{R}$ is minimized. This is typically done using a loss function $L_{e_n}(y_i, \hat{y}_i)$. The loss function quantifies the deviation between the true value and the prediction. In our case, the loss function is weighted by the proportion of samples belonging to the cluster associated with Expert $e_n$. This weighting factor is defined as:

$$\lambda_i = C_i[C_{i,max}] \in ]0,1] \; \forall i$$

This step significantly differs from the standard MoE as described in the background chapter. Without weighting the loss, an instance contributes equally to every Expert with a factor of 1, even though, as shown in the example from Figure 3, only a percentage $0 < \lambda_i \leq 1$ of the PDF falls within the Expert's region. When the number of Experts $|E|$ increases, $\lambda_i$ will most likely decrease and if the extreme case of $|E| \to \infty$ occurs, $\lambda_i$ normally approaches zero. Therefore, weighting is particularly crucial as it ensures that the loss of an instance is incorporated into the training based on the proportion of how much the instance belongs to the Expert's subspace. Let the loss for Expert $e_n$ be defined as:

$$L_{e_n}(y_i, \hat{y}_i) = \lambda_i \sum_{i=1}^{I} f_{loss}(y_i, \hat{y}_i)$$

where:

- $\lambda_i$ serves as weighting variable for the loss of $e_n$, which is defined as the probability, that instance $i$ lies within the region of $e_n$
- $f_{loss}$ represents the set of all suitable loss functions for $E$
- $y_i$ stands for the prediction of Expert $e_n$ for the instance $(m_{i,local}, x_{ik*})$
- $\hat{y}_i$ is the ground truth of the instance.

During training, the loss $L_{e_n}$ will be mimimized for $e_n$ through the gradient based backpropagation algorithm, which aims to adjust stepwise the weights $w_{t,t-1}^{(l)}$ and biases $b_t^{(l)}$ of the Expert.

$$\theta_{e_n}^*: \left(w_{e_n,t,t-1}^{(l)}, b_{e_n,t}^{(l)}\right)$$
$$\to argmin\left(L_{e_n}\left(y_i, \hat{y}_i\left(w_{e_n,t,t-1}^{(l)}, b_{e_n,t}^{(l)}\right)\right)\right)$$

$\theta_{e_n}^*$ represents for $\left(w_{e_n,t,t-1}^{(l)}, b_{e_n,t}^{(l)}\right) \forall t, l$, the parameters of $e_n$, which result in the lowest possible loss, after training with $r$ epochs by a given learningrate $\alpha \in \mathbb{R}^+$.

**Section 4: Training of Gating Unit with additional information**
After training the Experts, the final step involves training the Gating Unit, which takes the additional information of instance $i$ about the corresponding sample distribution along the subspaces as input to better weight the Experts under uncertainty. Let $D_{G,train}$ be the training dataset for the Gating Unit $G$, which consists of the global mode values $m_{i,global}$ representing the uncertain attributes, the certain attributes data $x_{ik*}$, the cluster probability vector $C_i$ as additional information for the distribution of uncertainty across the subspaces and the corresponding predictions of the Experts of every instance $\hat{y}_{i,e_n}$ in the training dataset:

$$D_{Gate,train} = \left\{\left((m_{i,global}, x_{ik*}), C_i, \hat{y}_{i,e_n}\right)\right\} \forall i, n$$

As output of the Gating Unit the weights $G_{e_n} = \{g_{e_1}, g_{e_2}, \ldots, g_{e_N}\}$ for the Experts are generated by a softmax activation in the last layer $L$ of the Gating Unit:

$$g_{e_n} = \frac{exp(z_t^L)}{\sum_{t=1}^{|E|} exp(z_t^L)}, \quad g_{e_n} \in [0,1]$$

where:

- $|E|$: Total number of Experts
- $z_t^L$: Is the weighted sum of inputs for the $t$-th neuron in the $L$-th layer

The training process of the Gating Unit involves learning the weightings $g_{e_n}$ to minimize the loss function of the uMoE model including the Gating Unit and the Experts. This is typically achieved through backpropagation and gradient descent. At first the weighting components are initialized randomly and can be seen as weights, similar to the NN parameters. During training, the Gating Unit learns the patters in the data, to weight the output of the Experts for each instance to reduce the total loss $L_{uMoE}$, which can be described as following:

$$L_{uMoE}(y_i, \hat{y}_{i,e_n}, g_{i,e_n}) = -\sum_{n=1}^{|E|}\sum_{i=1}^{I} y_i \log(g_{i,e_n} \hat{y}_{i,e_n})$$

Like the backpropagation of the Experts, the parameters of the Gating Unit $\theta_G^* = (w_{G,t,t-1}^{(l)}, b_{G,t}^{(l)})$ are optimized to minimize the loss $L_{uMoE}$. Thus, the learning process of the parameters can be expressed as:

$$\theta_G^*: (\theta_{e_n}^*, \theta_G)$$
$$\to argmin\left(L_{MoE}\left(y_i, \hat{y}_{i,e_n}(\theta_{e_n}^*), g_{i,e_n}(\theta_G)\right)\right)$$

where:
- $\theta_{e_n}^*$ : Trained parameters of Expert $e_n$
- $\theta_G$ : Parameters of the Gating Unit
- $g_{i,e_n}(\theta_G)$ : The weight of the Gating Unit for the Expert $e_n$ given the parameters $\theta_G$

**Section 5: Inference on trained uMoE model**
After the parameters of the clustering object $\theta_{kmeans}^*$, the individual Experts $\theta_{e_n}^*$ and the Gating Unit $\theta_G^*$ have been sequentially trained to minimize the loss function, the uMoE $\theta_{uMoE}^* = \{\theta_{kmeans}^*, \theta_{e_n}^*, \theta_G^*\}$ model is ready for inference on new instances, both those with instances that contain only certain attributes ($u = 0$, $0 \leq u \leq 1$), as well as for instances that contain at least one uncertain attribute ($u > 0$). In this context, $u$ is utilized as an indicator measuring the average percentage of uncertain attributes in an instance.

$$\hat{y}_i(u) = \begin{cases} f_{uMoE}: (x_{ik}) \to \mathbb{R}, & u = 0 \\ f_{uMoE}: (pdf_i) \to \mathbb{R}, & u > 0 \end{cases}$$

This trained model can now be employed to make predictions and perform inference tasks, leveraging the collective expertise of its constituent Experts and the learned gating mechanism. For instances with known attributes ($u = 0$), the inference process is straightforward. The model computes the prediction $\hat{y}_i$ by passing the instance $x_i$ through the gating mechanism and the corresponding Expert.

In this case, the additional information used by the gating mechanism is essentially $C_i$ as one-hot encoding vector $C_i(kmeans(x_{ik})) = [0,0,\ldots,0,1,0,\ldots,0]$, indicating the assignment of the instance to a specific cluster using the trained k-means object.

For instances with uncertain attributes ($u > 0$), the inference process involves an additional step. Since the attributes are uncertain, the model samples $S_i = \{s_{i,1}, s_{i,2}, \ldots, s_{i,M}\}, s \in \mathbb{R}^{k'}$ from the PDF ($pdf_i$) associated with the instance to generate the additional variables that characterize the cluster distribution. The additional variables are then provided to the gating mechanism, including the global mode value of the PDF.

Within the scope of this work, the primary focus has been on addressing uncertainty in the input space during training. As a result, when it comes to the inference phase for evaluation purposes, we will primarily consider scenarios with certain data, where $u = 0$. However, it is worth noting, that our model possesses the capability to handle uncertain instances even when working with the trained model. While this aspect is not the central point of evaluation in this context, it underscores the model's ability to deal with uncertainty in both training and inference phases.

**Section 6: Nested Cross-Validation for hyperparameter-tuning of number of subspaces**
After presenting the individual steps from subspace partitioning to training the components of the uMoE model and the inference process in sections 1-5, section 6 serves as presentation of an algorithm to evaluate the optimal number of subspaces. As mentioned in section 1, the partitioning of the input space and the resulting number of subspaces play a crucial role. However, since we must consider the partitioning of the input space under uncertainty, common pre-deterministic methods for determining the optimal number of clusters, such as the Elbow Method, are not applicable. This is because, due to the uncertainty and associated variance in the data, the data points are spread much wider across the input space. With the classical elbow method, it is no longer possible to clearly discern cluster separation on this scattered data, as each data point is treated as an instance and does not contribute as a weighted sample from an instance into the model, unlike in our approach.

For this reason, our uMoE approach is embedded in a Nested Cross-Validation (NCV), which allows us to determine the number of subspaces, and thus the cluster size, as well as the associated number of Experts for each dataset. In general, NCV consists of two loops, with the inner loop attempting to determine the optimal hyperparameters for the training data in the outer loop by optimizing hyperparameters and then applying them subsequently to the training data in the outer loop (Krstajic et al. 2014; Bates et al. 2023). In the outer loop, the dataset is partitioned into $a$ subsets, known as outer folds $D_{train}^a$. One of these outer folds is designated as the test set, while the remaining $a -$

1 folds collectively form the training set. For each outer fold, the following steps are performed:

Within each outer fold, the training set $D_{train}^a$ is further partitioned into $b$ subsets, known as inner folds. One of these inner folds serves as the validation set $D_{b,val}^a$, while the remaining $b-1$ subsets are used for hyperparameter tuning. For each inner fold, the uMoE model is trained and evaluated for different number of subspaces. The optimal number of subspaces $n_{subpace}^* \in \mathbb{Z}^+ \setminus \{1\}$, which archives the lowest loss on the validation set will be used for training and testing on the outer fold. For each $n_{subpace}^*$, the uMoE model is trained using the entire training data of the corresponding outer fold. Subsequently, the model's performance is assessed on the test set of the respective outer fold. The process of selecting the optimal number of subspaces is repeated for each outer fold, resulting in the set of $n_{subspace}^*$ for each outer loop $N = \{n_{subspace,a=1}^*, n_{subspace,a=2}^*, \ldots, n_{subspace,a=A}^*\}$.

## Practical Instructions for uMoE

The uMoE model is supported by an extensive GitHub repository, encompassing all essential resources and implementations. This repository has been curated to facilitate seamless and effective utilization of the uMoE model across various applications. The link to the code is attached to the abstract.

This section offers a concise depiction of what can be encounter within the repository. The uMoE streamlines the entire process of a basic Machine Learning pipeline, offering essential functionalities for training, prediction, and model assessment.

**Preprocessing uncertain instances**

In the real-world context, working with datasets containing uncertain or missing instances is a common challenge. To facilitate the handling of such uncertain data, the project incorporates the uframe library (Christian Amesoeder and Michael Hagn 2023). This extension enables the generation of highly complex multidimensional PDFs for uncertain instances using Multiple Imputations by Chained Equations – MICE (Buuren and Groothuis-Oudshoorn 2011). It also transforms instances into objects, granting access to a range of functions, including global mode calculation and sampling. Due to its numerous useful and accessible features for simplifying the handling of uncertain data, we have built our approach on this library. Therefore, as start, the uMoE model expects training data in the form of uframe objects. In a real-world environment, uncertain attributes of an instance can be manually deleted, thereby marking them as uncertain for uframe. Subsequently, uframe generates a PDF using MICE and stores it as an object. In academic settings, attributes can be marked as uncertain at will, serving as the training foundation for uMoE.

**Train uMoE object with fit()**

After the data has been preprocessed and converted into uframe objects, an initialized uMoE object can be trained on the uncertain instances using the fit() function. This highly customizable function allows for the specification of various hyperparameters, including the learning rate, the number of training epochs, batch sizes for both Experts and the Gating Unit as well as their corresponding NN-Architectures (number of hidden layers and neurons per layer). This adaptability ensures the model's suitability for diverse datasets and tasks. To prevent overfitting, the fit method leverages Elastic Net regularization (Zou and Hastie 2005), which combines L1 (Lasso) and L2 (Ridge) regularization techniques. Elastic Net encourages sparse model weights while also handling highly correlated features effectively. This regularization approach helps maintain model robustness, even in the presence of noisy or uncertain data. The user can also set a threshold parameter $p$, which controls how much of the uncertainty of the instance the model experiences (see Section 1). By choosing $p = 1$ the hole PDF is used for training, while values near 0 assume that only a small area around the global mode value is considered. For the determination of the local and global mode values as described in section 2, the uMoE method includes the basin hopping optimizer, due to the ability of finding the maxima in a highly-dimensional landscape, what agrees with a multidimensional PDF (Wales and Doye 1997).

**Prediction with predict()**

After training a uMoE model, the predict method comes into play. This function is designed to make predictions on certain instances (see section 5). It employs clustering to assign instances to the relevant Experts and then utilizes these Experts to make predictions. The cluster probabilities for each instance are than further used additional input information for the Gating Unit. The final predictions of the uMoE are returned in a list.

**Evaluation with evaluation()**

The evaluation function calculates the performance of the uMoE model's predictions against the true target values. It generates scores tailored to the specific task type. For classification tasks, it computes the accuracy score, while for regression tasks, it calculates the mean squared error. This function is crucial for assessing how well the model's predictions align with the ground truth.

## Evaluation

In this section, we conduct a comprehensive evaluation of the uMoE model. We begin by describing our evaluation procedure, which firstly encompasses the generation of uncertainty, a short presentation of the used datasets, as well as a description of the associated references. Subsequently, we introduce two evaluation perspectives. The first assesses our model's ability to learn from uncertain data and test on certain data using NCV for subspace size determination. The second perspective involves iterative evaluation over increasing numbers of subspaces to visualize the performance graphically.

We begin by discussing the topic of uncertainty in datasets. As we were unable to find suitable datasets with aleatoric uncertainty, we introduced uncertainty artificially by randomly removing $u$ percent of the data for each dataset. We then generated one PDF for each uncertain instance using uframe and MICE. In our context, MICE serves as intermediate step to create PDFs through generating multiple possible imputation values for all missing data points. Following this, the imputations can then be fused into one multidimensional PDF across all attributes with a bandwidth of 0.1 as smoothing parameter of the kernel density estimation. It is important to note that the ground truths corresponding to each uncertain instance remained unchanged. We assumed that these ground truths would not fundamentally change due to uncertainty, and our focus was solely on uncertainty within the input space.

### Datasets

Having introduced the generation of uncertainty, the next step is to provide an overview of all the datasets we considered for evaluation. For each dataset, we have summarized the source reference, the number of attributes and instances as well as the prediction task (Regression (**R**) and Classification (**C**)). When selecting the datasets, we exclusively focused on well-known online databases such as Kaggle and the UC Irvine Machine Learning Repository.

| Dataset | Reference | # Attributes | # Instances | Task |
|---|---|---|---|---|
| Blood | (I-Cheng Yeh 2008) | 4 | 748 | C |
| California Housing | (Kelley Pace and Barry 1997) | 8 | 20600 | R |
| Energy Efficiency | (Tsanas and Xifara 2012) | 8 | 768 | R |
| Water Potability | (Aditya 2021) | 9 | 2011 | C |
| Diabetes | (Smith et al. 1988) | 8 | 768 | C |
| Banana | (Jaichandaran 2023) | 2 | 5300 | C |
| Red Wine | (Cortez et al. 2009) | 11 | 1599 | C |

Table 1: Summary of datasets included in the evaluation

### Reference Methods

To better assess the success of our method, additional reference approaches were incorporated. As previously described in the related work chapter, there are only a few works that have dealt with training with uncertain input data. However, when considering comparisons to our research, no existing method serves as a suitable reference. Ren et al.'s approach is confined to Naïve Bayes limits and classification prediction tasks only like the decision tree approach of Tsang et al.. Kendall and Gal's method is likewise unsuitable because it focuses on image data. Gast and Roth are also unsuitable as references since their method incorporates inference with aleatoric uncertainty. For this reason, specific baseline references were chosen, capable of working with PDFs and simultaneously are competitive with our method. We selected four references, all of which fall in the field of NNs. These four approaches can be separated into two groups: The first group was trained using the global mode values $m_{global}$ of the PDFs, and the other using the expected values $\mu$ of the PDFs, effectively covering a broad spectrum of deterministic measures within a PDF. Both deterministic values were then assigned to a standard NN, as well as to a classical MoE, which leads to totally four independently approaches. Concretely speaking, for the NN references a model was trained with either the global mode values or the expected values and subsequently tested on certain, unseen data. In the case of the MoE references, the expected value or the mode value was also provided as input, which was then clustered and assigned to an Expert. After each Expert received an instance assignment, they were trained accordingly, followed by the training of the Gating Unit (though this was done without the developed approaches from section 1-4, such as additional information or loss function weighting).

### Experimental Result 1: Mean Value of NCV

In this section, we will first introduce how we initialized the NN architectures of our uMoE components and the references. In conclusion we explain our evaluation approach trough NCV and show the results in tabular form.

For the reference methods as well as for the components of our uMoE, NNs with two hidden layers, each consisting of 16 neurons and using the ReLU activation function, were chosen. The learning rate during parameter training was set moderately at 0.01, with an epoch count of 150 and a batch size of 16 for both the reference NN and the Expert components of uMoE and MoE. The batch size for the Gating Unit is 24. Regarding regularization, alpha in the Elastic Net was set to 0.5, indicating that both L1 and L2 regularization have equal influence on the loss, with a lambda (regularization strength) of 0.002. This parameter setting for the NNs, both for the baselines and for the uMoE, was determined empirically through evaluations on the datasets from Table 1 and, in total, yielded the best training results for each method.

|  | Dataset | % Uncertainty ($u$) | uMoE | Ref. MoE ($m_{global}$) | Ref. MoE ($\mu$) | NN ($m_{global}$) | NN ($\mu$) |
|---|---|---|---|---|---|---|---|
| **Regression (MSE)** | California | 40 | 0.53 | 0.56 | 0.55 | **0.51** | 0.55 |
|  |  | 60 | **0.66** | 0.70 | 0.74 | 0.71 | 0.72 |
|  | Energy | 40 | **13.6** | 14.9 | 15.3 | 16.8 | 18.6 |
|  |  | 60 | 19 | 17.6 | **17.4** | 46.7 | 35.0 |
| **Classification (AUC)** | Diabetes | 40 | **75.3** | 72.8 | 71.3 | 74.0 | 73.2 |
|  |  | 60 | **72.1** | 70.3 | 69.8 | 68.0 | 67.8 |
|  | Blood | 40 | **76.6** | 76.2 | 76.2 | 76.2 | 76.2 |
|  |  | 60 | **76.4** | 75.7 | 76.0 | 76.0 | 76.2 |
|  | Banana | 40 | **82.1** | 80.7 | 80.9 | 73.4 | 75.2 |
|  |  | 60 | 58.3 | 58.3 | **62.5** | 58.3 | 58.3 |
|  | Water Potability | 40 | **63.1** | 62.0 | 60.8 | 60.5 | 61.7 |
|  |  | 60 | **59.5** | 57.5 | 55.8 | 57.5 | 56.8 |
|  | Wine Quality | 40 | **57.8** | 56.1 | 56.7 | **57.8** | **57.8** |
|  |  | 60 | **55.0** | 51.4 | 52.7 | 52.5 | 47.7 |

Table 2: Tabular evaluation with NCV

In addition to the general hyperparameters for the NNs, we selected a value of 0.8 for the threshold parameter $p$ in our method, which represents the threshold for the samples, at $u = 0.4$. As the uncertainty increased to $u = 0.6$, we reduced this threshold value from 0.8 to $p = 0.6$. In general, it can be said that as uncertainty increases, a low threshold is advantageous, and vice versa.

The results presented in Table 2 were generated by choosing the size $a = 5$ (outer fold - validation) and $b = 3$ (inner fold – subspace tuning) for the NCV for both our uMoE and the reference MoE. This process aimed to identify the optimal number of subspaces for each approach. Subsequently, the mean value across all the results of the outer folds was calculated. The references employing a baseline NN were each trained and tested on the same outer folds, followed by mean value calculation. The results were subsequently compiled into Table 2 and for each dataset the dominant values were highlighted. From the results in Table 1, it can be observed that our method predominantly outperforms the baseline methods. Considering the basic MoE as a part of our conclusion and then comparing it to the baseline NNs, our method performs only poorly in one out of 14 evaluations, with the deviations in these cases being very minimal.

**Experimental Result 2: Subspace Iteration**

To provide a deeper insight into the performance of the uMoE across varying number of subspaces, we conducted a second evaluation on a subset of the previous datasets. Unlike the previous tabular analysis focused on finding the ideal number of subspaces and evaluating their performance, this evaluation aims to demonstrate how uMoE performances compared to the references as the number of subspaces increases in a predefined interval.

For this evaluation, we conducted a basic Cross-Validation (without inner fold) for each dataset once again, to observe how performance varies across the number of subspaces (= number of Experts). As a result, the outcomes may differ from those in the tabular analysis. The interval was set in the range of two to six. The NNs, which were trained on the global mode values or expected values were tested on the same folds. Graphically the basic NN references are represented as vertical lines. The specific associations can be found in the legends within the plots. In the graphical evaluation, it becomes evident that our method is consistently better than the references, but the optimum is mostly located within a smaller subspace, typically ranging from two to four, and worsens as the number increases.

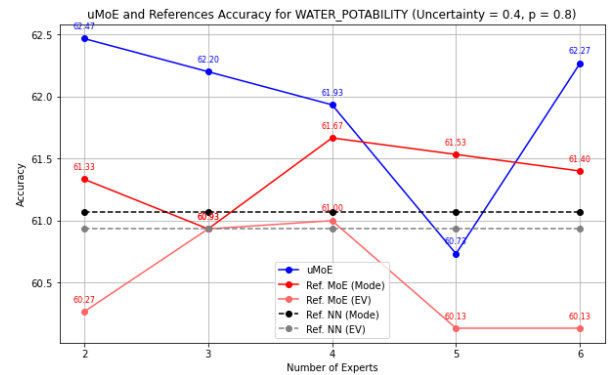

Figure 4: Subspace evaluation of Water Potability with $u = 0.4$ and $p = 0.8$

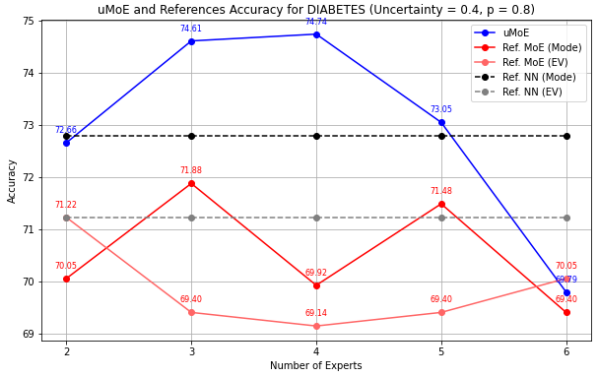

**Figure 5:** Subspace evaluation of Diabetes with $u = 0.4$ and $p = 0.8$

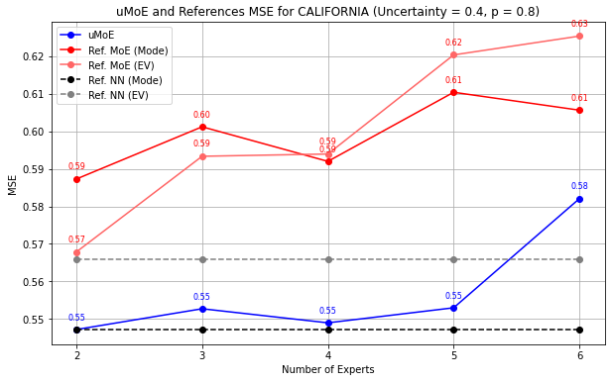

**Figure 6:** Subspace evaluation of California Housing with $u = 0.4$ and $p = 0.8$

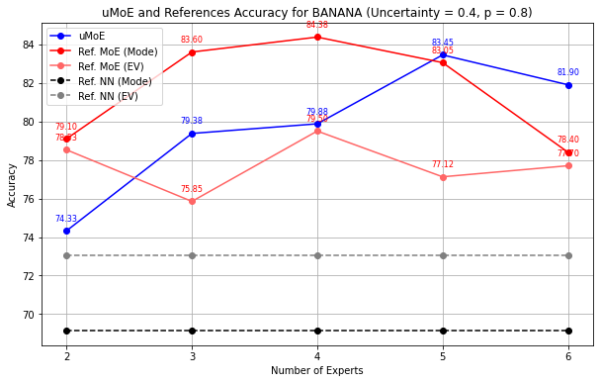

**Figure 7:** Subspace evaluation of Banana with $u = 0.4$ and $p = 0.8$

## Discussion

In this chapter, our objective is to provide an interpretation and critical discussion of the results obtained from the previous evaluation approaches. Within this context, we also performed a robustness analysis by iteratively selecting various sample threshold values, denoted as $p$. Besides that, we also intend to explain the limitations of our approach and show interesting future research directions based on our findings. From the tabular evaluation, it is evident that, with few exceptions, our approach emerged as the dominant method. The learned weights of the Gating Unit, influenced by additional information, and weighted training of the Experts with local mode values contributes to the model's enhanced consideration of uncertainty in the parameter optimization. While our method faced more pronounced competition in two cases, it was notably outperformed only by the standard MoE methods, which we consider as an integral component of our approach. With one rare exception, our model consistently outperformed the NN baseline methods, aligning with the primary objective of our evaluation.

The reason, why our method cannot always be dominant lies in the way how the uncertainty is represented in the PDFs. When MICE imputation accurately determines the global mode or expected values close to the true attribute values, our model tends to generalize too much based on the local mode values. This effect is amplified when also the variance of the PDF increases, leading to more cases where most of the probability mass is in a different subspace, than the true value would lies. It is precisely for such cases that the threshold for the samples becomes relevant, as it provides our model with flexibility to mitigate the variance by artificially constraining it and thereby reducing the search space for the local mode value.

This is one of the main disadvantages of subspace decomposition, that although leading to a value in the subspace of the Expert, the deviation from the real value can become significantly, especially with increasing number of subspaces. Moreover, it is evident that an increase in uncertainty results in a widening gap between uMoE and NN methods besides a few exceptional cases. This assertion underlines our theory that with more uncertainty the expected values and mode values deviate more and more clearly from the true value and thus the basline models fail to find the optimal parameters for inference on certain values. In summary, the tabular evaluation underscores that our uMoE significantly leads to better results through incorporating uncertainty in the training process.

In addition to the tabular evaluation, the graphical evaluation provides valuable insights into the number of subspaces necessary. Firstly, it is noticeable that the ideal subspace size for our uMoE varies from one dataset to another, yet the optimum is typically achieved at a lower subspace size, typically ranging from two to four. Furthermore, it is evident that performance generally decreases as the number of subspaces increases, with some exceptions that can be considered as outliers. This decline in performance can be attributed to the decreasing portion of uncertain instance objects allocated to each subspace as the number of subspaces expands. As a result, the decisions related to subspace cor-respondence and thus the locale mode search become increasingly restricted, as discussed before. Although this effect is partially mitigated by the weighting of the loss function, it cannot be fully addressed in the determination of local modes. Conclusively, it becomes evident that our model demonstrates the capability to adapt to uncertainty even with an inappropriate number of subspaces, showcasing its flexibility in responding to hyperparameters.

Having demonstrated the effectiveness of our method in the evaluation procedures, we now turn our attention to assessing the robustness of our approach concerning the threshold parameter $p$. To conduct this evaluation, we utilized the California Housing dataset. This choice was motivated by the fact that even slight changes in a regression dataset can have an immediate and noticeable impact on the target variable, unlike a classification dataset. We performed the evaluation for both $u = 0.4$ and $u = 0.6$ using varying threshold values ranging from 1 to 0.1 in increments of 0.1. This analysis was conducted through a NCV with $a = 3$ and $b = 2$. The result was then calculated using the mean value over all outer folds $a$ like in the tabular evaluation.

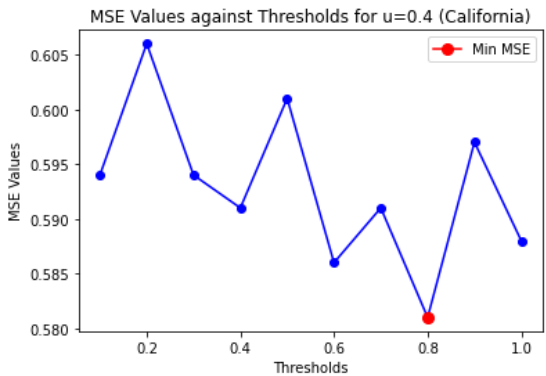

Figure 8: Sample threshold analysis for $u = 0.4$

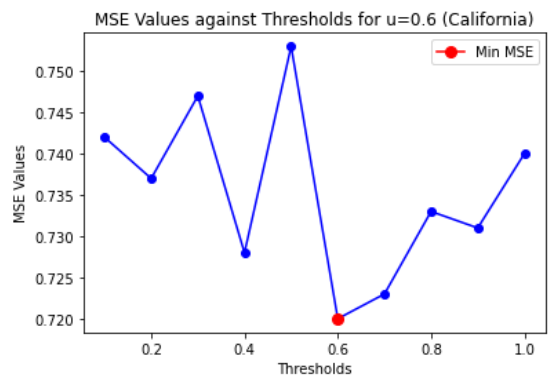

Figure 9: Sample threshold analysis for $u = 0.6$

Based on the outcomes, we conclude that in scenarios with low uncertainty, it is advisable to choose a higher $p$ value (e.g., $u = 0.4$ corresponds to $p = 0.8$), while in situations with high uncertainty, a lower $p$ value is more appropriate (e.g., $u = 0.6$ corresponds to $p = 0.6$). Furthermore, it is worth noting that fluctuations in the results between the different threshold values become minimal once $p$ values exceed 0.5. These fluctuations can be attributed to the underlaying NN components, which suffer from a non-convex optimization problem, due to stochastic gradient descent and thus can stuck in different local minima. Apart from this, the trend can already be derived that with increasing uncertainty, a lower threshold is more suitable despite small fluctuations. Consequently, we pre-determined the $p$ values for the previous evaluations (see Experimental Results 1 and 2) based on these findings. Incorporating both the $p$ value and the number of subspaces into the NCV would result in an excessively large number of possible combinations. Even if the influence of the threshold value is relatively small, it is still important, due to enabling our method to respond more robustly to increasing uncertainty levels. This adjustment is also important, because excessive uncertainty can lead to overgeneralization by the Experts when dealing with uncertain data. This, in turn, prevents the subgroups on which the Experts focus from becoming too large as already assumed in the previous part of the discussion. However, it is important to note that this reduction comes at the cost of losing some information about the distribution of the PDF.

Having discussed the ability of our method to dominate the baseline methods and to respond to increasing variance in PDFs, it is important to also acknowledge the limitations our model in the following paragraph. One limitation pertains to the fact that our model relies on numerous hyperparameters, whose predetermination is still a research subject. These encompass typical hyperparameters of the NN architecture, which are integral to our uMoE, such as the number of hidden layers, neurons, and so forth. Additionally, this includes the selection of the threshold value and, notably, the choice of the number of subspaces in our case. While this crucial hyperparameter can be determined using NCV, it is dependent on the size of the inner folds, rendering it time intensive. This aspect may not be particularly relevant during the training process but still necessitates consideration. Furthermore, it is worth mentioning that while our method accounts for uncertainty in the input when optimizing parameters, it does not quantify this uncertainty in the output during inference on certain instances. Consequently, the user must place trust in the model to provide the best possible prediction without furnishing information regarding the uncertainty in the output.

Towards the conclusion of this chapter, we aim to discuss potential avenues for future research. One intriguing research direction involves delving deeper into the selection of predictive models for the individual components of the uMoE. In this regard, emerging large-scale data models could gain significance, especially those trained on extensive data under uncertainty, subsequently serving as components in the Gating Unit and/or Expert components. This approach could also be extended to the choice of clustering algorithms. Beyond model selection, it holds promise to explore the inference process of a trained uMoE in greater detail, particularly in propagating uncertainty through the trained uMoE for more precise handling of uncertain data. These aspects offer potential for future extensions of the uMoE method.

## Conclusion

We have explored uncertainty in data, particularly aleatoric uncertainty, and its impact on training of NNs. Existing research has mainly focused on uncertainty during inference, without paying attention to the challenge that arises when the data used for training is generated under uncertainty. To address this gap, we introduced the uMoE method, which follows the "Divide and Conquer" paradigm. This paradigm allows our model to break down the complex problem of uncertainty represented as continuous PDFs into smaller subspaces and additionally using the obtained information from the decomposition to find the optimal parameters, that represent the uncertainty in the data best. Through extensive evaluations, we demonstrated uMoE's effectiveness, outperforming baseline methods in learning from uncertain data and then use the trained models to predict on certain data.

This outstanding performance is based on our method's ability to respond to different amount of variance in the PDFs by partitioning them into subspaces. More detailed said, our method locates the local mode of the PDF within a subspace and integrates it in weighted manner into the training of so-called Expert models responsible for that subspace. The Gating Unit on top of the Experts subsequently learns to weight the predictions of the Expert models with gained additional information about the distribution of the PDFs across the subspaces. As result, with each training iteration, the additional information from the decomposition about the variance of the PDF is iteratively incorporated into multiple training steps as described in the "Divide and Conquer" paradigm. It is also important to note that our method has no restrictions on the input PDFs and can handle even non-parametric distributions.

In summary, our work advances uncertainty-aware Deep Learning during training, offering practical instructions for implementing the uMoE. We stress the importance of integrating uncertainty awareness into predictive models in general to enhance the reliability of AI systems. Our achievements open doors for future research in this field, promising to enhance AI system capabilities and trustworthiness.